\documentclass{article}
\usepackage{amssymb}
\usepackage{multirow}
\usepackage{booktabs}
\usepackage[dvipsnames, svgnames, x11names, table]{xcolor}  % ✅ 只加载一次，包含所有需要的选项
\usepackage{amsmath} 
\usepackage{colortbl}
\usepackage{graphicx}
\usepackage{caption}
\usepackage{array}
\usepackage{listings}
\usepackage{subcaption}

\definecolor{deepPink}{RGB}{255, 20, 147}
\usepackage[preprint]{corl_2025} % Uncomment for pre-prints (e.g., arxiv); This is like ``final'', but will remove the CORL footnote.

\title{TypeTele: Releasing Dexterity in Teleoperation \\ by Dexterous Manipulation Types}

\author{
  Yuhao Lin\footnotemark[1] \textsuperscript{ 1}, 
  Yi-Lin Wei\footnotemark[1] \textsuperscript{ 1}, 
  Haoran Liao\textsuperscript{ 1}, 
  Mu Lin\textsuperscript{ 1}, 
  Chengyi Xing\textsuperscript{ 2}, 
  Hao Li\textsuperscript{ 2} \\ 
  \textbf{Dandan Zhang\textsuperscript{ 3},} 
  \textbf{Mark Cutkosky\textsuperscript{ 2},}
  \textbf{Wei-Shi Zheng\textsuperscript{ 1}} \\
  \textsuperscript{1} School of Computer Science and Engineering, Sun Yat-sen University, China \\
  \textsuperscript{2} Stanford University, USA, \textsuperscript{3} Imperial College London, UK \\
  \href{https://isee-laboratory.github.io/TypeTele/}{\textcolor{deepPink}{https://isee-laboratory.github.io/TypeTele}}
}

\begin{document}
\maketitle
\vspace{-3em}
\begin{figure}[h]
    \centering
    \includegraphics[width=\linewidth]{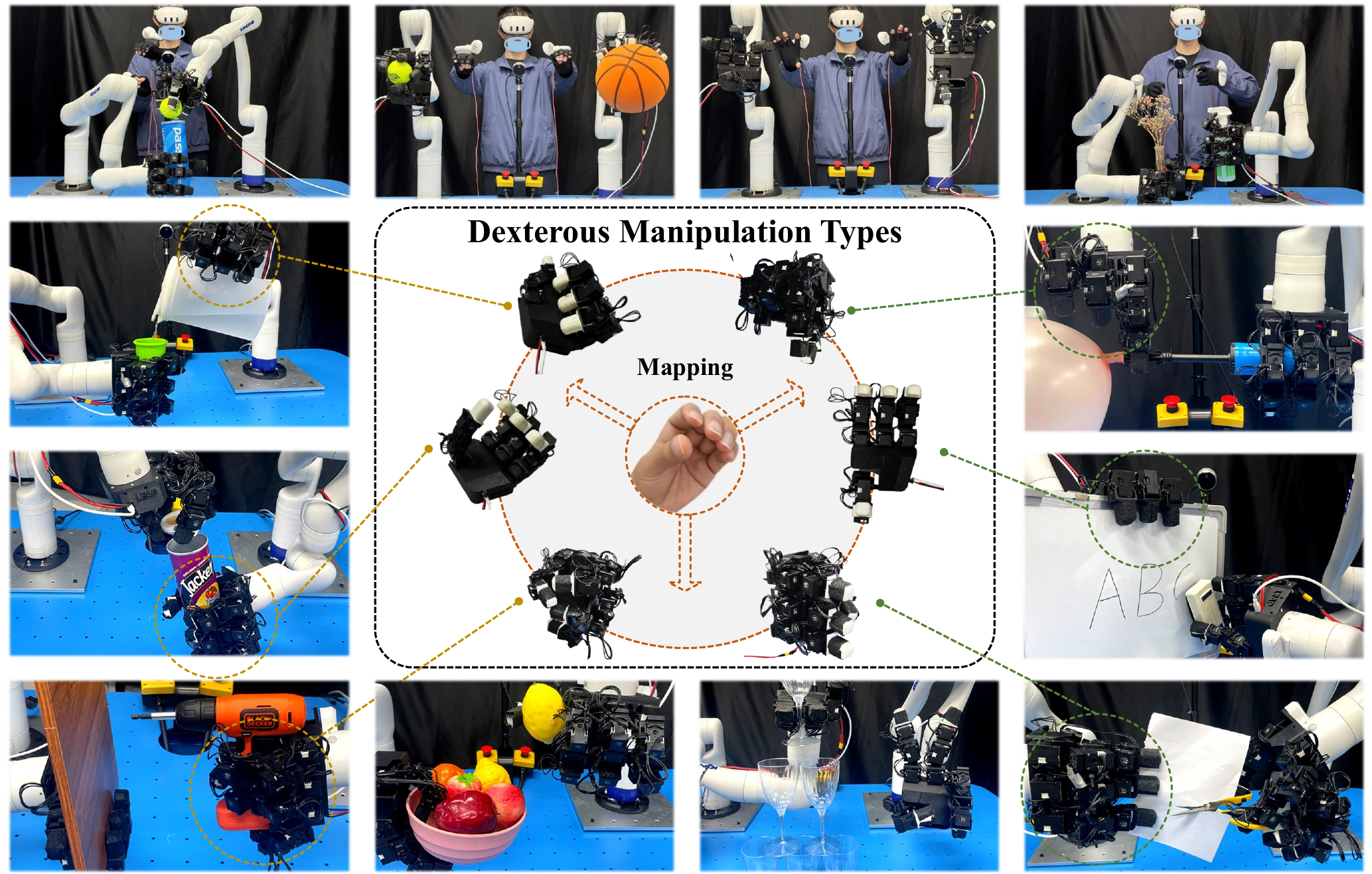}
    \caption{
    \textbf{TypeTele}, an effective dexterous teleoperation system, enables operators to complete various manipulation tasks by corresponding human hands with different types of robotic hands.
    }
    \label{fig:setting}
\vspace{-1em}
\end{figure}
%===============================================================================

\begin{abstract}
Dexterous teleoperation plays a crucial role in robotic manipulation for real-world data collection and remote robot control. Previous dexterous teleoperation mostly relies on hand retargeting to closely mimic human hand postures. However, these approaches may fail to fully leverage the inherent dexterity of dexterous hands, which can execute unique actions through their structural advantages compared to human hands. To address this limitation, we propose \textbf{TypeTele}, a type-guided dexterous teleoperation system, which enables dexterous hands to perform actions that are not constrained by human motion patterns. This is achieved by introducing dexterous manipulation types into the teleoperation system, allowing operators to employ appropriate types to complete specific tasks. To support this system, we build an extensible dexterous manipulation type library to cover comprehensive dexterous postures used in manipulation tasks. During teleoperation, we employ a MLLM (Multi-modality Large Language Model)-assisted type retrieval module to identify the most suitable manipulation type based on the specific task and operator commands. Extensive experiments of real-world teleoperation and imitation learning demonstrate that the incorporation of manipulation types significantly takes full advantage of the dexterous robot's ability to perform diverse and complex tasks with higher success rates.

\end{abstract}

%===============================================================================
\section{Introduction}

With the development of learning-based methods and large-scale robotic datasets, dexterous robots have become increasingly capable of performing diverse and delicate tasks \cite{brohan2023rt2, fang2025anydexgrasp}. Teleoperation plays a critical role in collecting real-world data, as it enables the acquisition of high-quality robotic demonstrations under realistic observations and physically executable actions \cite{Chen2024BunnyVisionPro, fu2025mobile_ALOHA, shawbimanualtele}.

Previous dexterous teleoperation methods aim to control the dexterous hand by imitating human hand postures, typically achieved by first capturing the human hand poses and retargeting them to the dexterous hand \cite{cheng2024Open-TeleVision, qin2023anyteleop, chen2024arcap, shawbimanualtele}. Most hand retargeting approaches employ optimization or inverse dynamics techniques to preserve the spatial consistency of vectors between the wrist and predefined keypoints (such as fingertips) in both the human and robotic hands \cite{shawbimanualtele, qin2023anyteleop, qin2022dexmv}. However, the retargeting paradigm is unable to fully utilize this dexterity of the dexterous hand, leading to difficulty in performing both basic and complex tasks.

\begin{figure}[t]
    % \vspace{-1em}
    \centering
    \includegraphics[width=\linewidth]{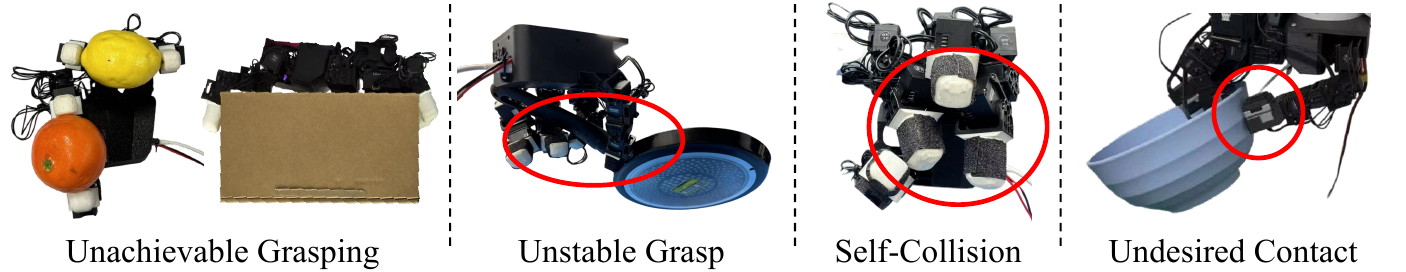}
    \caption{
    The challenges of previous retargeting-based dexterous teleoperation systems. Unachievable Grasping shows poses that are physically infeasible for human hands. Unstable Grasp leads to object dropping due to weak contact. Self-Collision indicates finger interference during motion. Undesired Contact refers to insufficient contact between the tactile sensor surfaces and the object.
    }
    \label{fig:problem}
    % \vspace{-1em}
\end{figure}

However, two challenges hinder the effectiveness of teleoperation in existing methods. \textbf{First, the retargeting paradigm restricts the dexterous hand to motions feasible for human hands}, as it enforces consistency between human and robotic hand postures. The fully actuated dexterous hand can perform poses that humans cannot, but are more suitable to complete specific manipulation tasks, as shown in the left of Fig. \ref{fig:problem}. \textbf{Second, morphological differences between human and robotic hands may lead to the unreasonable retargeting poses.} Existing methods typically align corresponding vectors between the two hands and solve for a pose in the full joint space. However, differences in kinematics often lead to unstable postures, self-collisions, or undesirable contact directions in the robotic hand \cite{weigraspasyousay, li2025maniptrans, guo2024telephantom}, as shown in Fig.~\ref{fig:problem}.

To overcome these problems, we propose \textbf{TypeTele}, a type-guided dexterous teleoperation system, which allows operators to employ appropriate dexterous manipulation types to manipulate different objects and complete different tasks. The introduction of types offers two benefits, which address the two aforementioned bottlenecks: \textbf{1)} Introducing dexterous manipulation types enables robots to perform actions that the human hand cannot perform. \textbf{2)} Dividing dexterous actions into discrete types improves the effectiveness and rationality of the dexterous hand postures. 

To support our system, we construct a dexterous manipulation type library organized using a hierarchical taxonomy that covers typical actions required in manipulation tasks. Each manipulation type is annotated with corresponding stretching and contracting postures of the robotic hand, which determines the range of feasible actions that can be executed within this type. Based on this library, our teleoperation framework operates in two stages: type retrieval and action execution. For type retrieval, we propose an MLLM (Multi-modality Large Language Model) assisted type retrieval module that identifies the most appropriate manipulation type based on the current task. For action execution, we design an interpolation mapping strategy that maps the natural human hand action to the specific dexterous manipulation type, thereby enabling intuitive control of the robotic hand through human motion. 

The experimental results demonstrate the effectiveness of our teleoperation system: 1) Our system enables the successful execution of tasks that are unachievable using retargeting-based teleoperation system. 2) Our system significantly improves the data collection efficiency. 3) The data collected by our system shows higher quality, which benefits subsequent imitation learning and enhances the performance and robustness of autonomous policies. 4) The key insight of introducing type into teleoperation shows strong applicability to various tasks and can be applied to different systems.

\section{Related Works}
\subsection{Dexterous Teleoperation}
Teleoperation is a fundamental task for robotics\cite{darvish2023survey1, hokayem2006survey2}, as it not only enables remote operation of robots but also facilitates data collection for imitation learning \cite{zhao2023aloha, cheng2024Open-TeleVision, he2024omnih2o}. Research on teleoperation for two-finger gripper robots primarily focuses on arm control, such as master-slave systems~\cite{aloha22024aloha2, fu2025mobile_ALOHA} and VR devices \cite{erkhov2025viewvr}, achieving impressive performance. Compared to grippers, dexterous hands offer greater dexterity for fine-grained manipulation tasks, but they also introduce challenges in hand pose mapping due to morphological differences between human and robotic hands~\cite{yin2025dexteritygen, qin2024from_one_hand, iyer2024openteach, guo2024telephantom}. Previous methods focus on human hand pose capture and pose mapping to closely mimic human hand postures, typically involving different hardware setups for motion capture \cite{qin2023anyteleop,yang2024ace, Chen2024BunnyVisionPro, chen2024arcap, shawbimanualtele} while sharing fundamentally similar retargeting algorithms \cite{qin2024from_one_hand, qin2022dexmv}. These methods face two key limitations: (1) they are limited to actions feasible for human hands and (2) pose mapping remains challenging due to morphological differences, hindering fine manipulation tasks. In this paper, we propose a type-guided teleoperation system to address these limitations, enabling more complex manipulation tasks despite morphological mismatches.

\subsection{Dexterous Manipulation}
Achieving autonomous and generalizable dexterous manipulation is a long-term goal for robotics community\cite{turing2009Turin, brohan2023rt2}. With the development of deep learning, imitation learning methods have shown great promise to achieve this goal \cite{chi2023dp, fang2025anydexgrasp, li2025maniptrans}, with transformer-based \cite{zhao2023aloha, wang2024unigrasptransformer}, diffusion-based\cite{ze2024dp3, ze2024idp3, xue2025demogen} or Vision-Language-Action based architecture \cite{zhong2025dexgraspvla, brohan2023rt2}. However, the effectiveness of these methods largely depends on the quality and scale of expert demonstration data \cite{li2025How_to_Train_Your_Robots, liu2024rdt, mu2024robotwin}. To address this, our system achieves higher data quality and collection efficiency, which facilitate more effective imitation learning and improve the performance of autonomous policies.

\subsection{System Overview}
\begin{figure}[t]
    \centering
    \includegraphics[width=\linewidth]{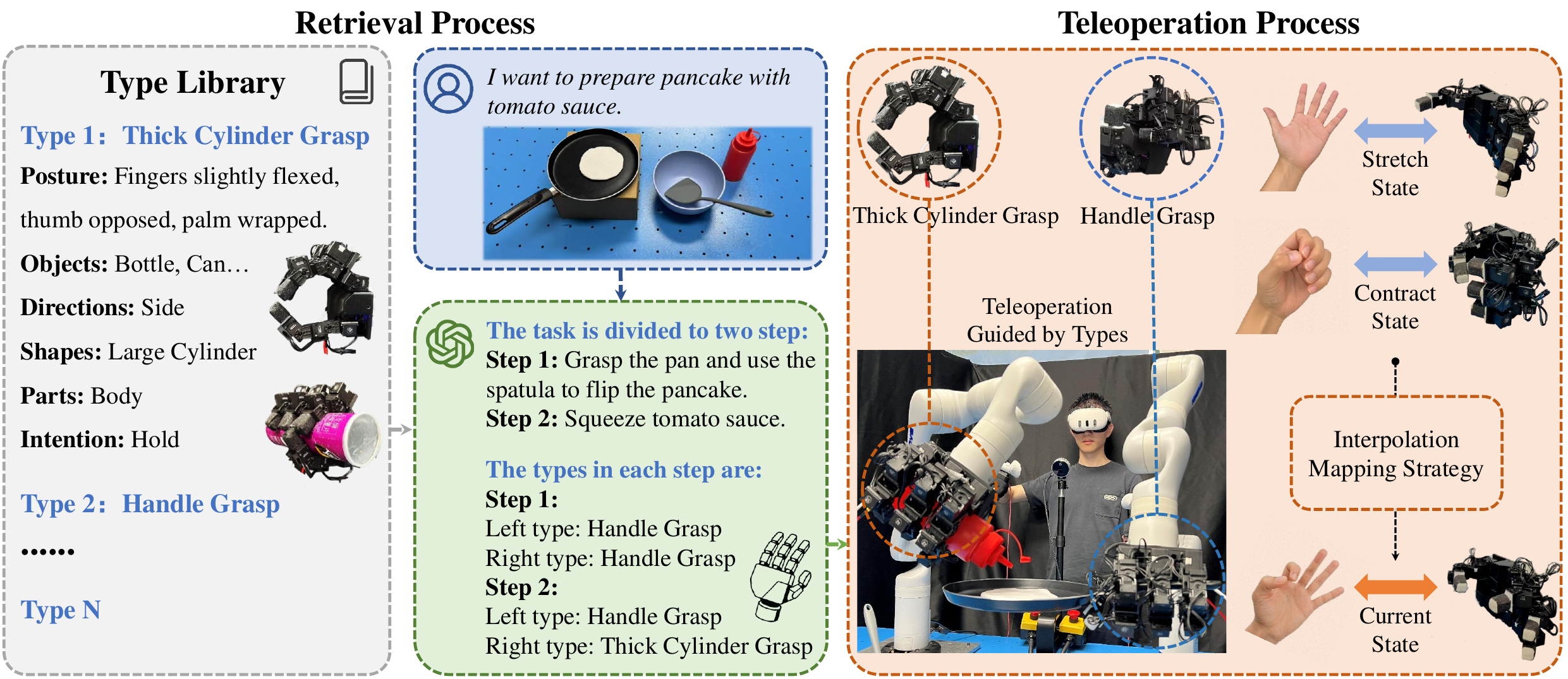}
    \caption{
    TypeTele includes a retrieval process using a MLLM to select manipulation types from the library, and a teleoperation process that applies them with an interpolation-based mapping strategy.
    }
    \label{fig:pipeline}
\vspace{-1em}
\end{figure}

The overview of our system is shown in Figure \ref{fig:pipeline}. First, we construct a dexterous manipulation type library, which covers the types required for various manipulation tasks. Then, we propose a MLLM-assisted type retrieval module to select the most appropriate manipulation type based on the current task. And we design a type adjustment strategy to improve its versatility. Finally, for the teleoperation process, we design an interpolation mapping strategy to control the dexterous action of specific type by human hand motion.

\section{Type-guided Teleoperation System}

\subsection{Dexterous Manipulation Type Library}
\begin{figure}[t]
    \centering
    \includegraphics[width=\linewidth]{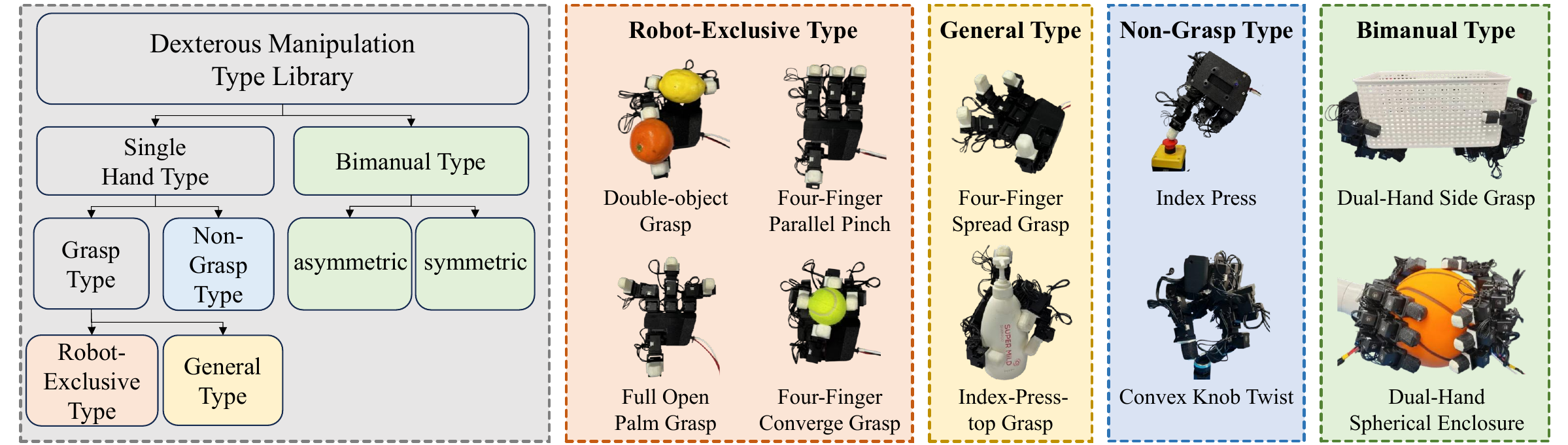}
    \caption{
    The illustration of the dexterous manipulation library. The left side presents the hierarchical taxonomy of the library, while the right side displays examples from each category.
    }
    \label{fig:typelibrary}
\end{figure}

Inspired by existing human grasp taxonomies that classify hand postures into distinct types to encompass most human manipulations~\cite{cutkosky1989grasptaxonomy2}, we design a \textbf{Dexterous Manipulation Type Library}, comprising diverse dexterous types to guide dexterous postures across a wide range of teleoperation tasks, as shown in Figure \ref{fig:typelibrary}. The library is built upon recent taxonomies \cite{feix2015grasptaxonomy1, cutkosky1989grasptaxonomy2, krebs2022bimanualtaxonomy, fang2025anydexgrasp, chen2024non-grasp} and augmented with postures specially designed for dexterous hands, which are extracted from a variety of dexterous manipulation tasks. 

To better organize the library and effectively cover the dexterous manipulation action space, we classify the dexterous manipulation types into two primary categories: Single-hand types and bimanual collaborative types, as illustrated in Figure \ref{fig:typelibrary}. Specifically, single-hand types are further subdivided into grasp types and non-grasp types. Among the grasp types, we define two subcategories: robot-exclusive grasp types, which support manipulation tasks that exceed the capabilities of human hands, and general grasp types, which are derived from established human grasp taxonomies. Furthermore, the bimanual collaborative types are subdivided into symmetric and asymmetric types based on the relative positions and functional roles of the two hands during manipulation. Overall, our library is consisted with 4 sub-categories and 30 types. 

Specifically, each dexterous type is annotated with stretching and contracting postures, which correspond to the natural stretching and contracting postures of the human hand. Additionally, to facilitate the autonomous retrieval of type, each dexterous type is annotated with object-centric and posture-centric information to describe: (1) what kinds of objects and task this posture is suitable for manipulating; (2) what the posture specifically looks like. These details are organized into manipulation attributes belonging to each dexterous type, as shown in the left of Figure \ref{fig:pipeline}, facilitating type retrieval in teleoperation. More details can be found in supplementary materials.

\subsection{MLLM-assisted Type Retrieval}

To facilitate teleoperation, we propose an MLLM-assisted manipulation type retrieval framework that autonomously selects the appropriate manipulation type for a given task. Specifically, all manipulation types in the library, annotated with attribute descriptions, are converted into language format prompts and fed into a MLLM, such as GPT-4o~\cite{hurst2024gpt}. We then prompt the MLLM to sequentially reason through two sub-questions: (1) \textit{How many steps are required to complete the task?} and (2) \textit{Which type of manipulation should be assigned to each hand at each step?} The MLLM is first guided to decompose the task into a series of steps, infer which objects are involved in each step, and determine the way of the interaction. Based on this reasoning, the MLLM infers the desired attributes of the manipulation type for each hand. These inferred attributes are then used to retrieve the most suitable manipulation type from the library.

We develop a voice control program to facilitate remote operators' interaction with the retrieval module, especially when their hands are occupied with robotic arm control tasks. This system utilizes Whisper~\cite{radford2023robust} to transcribe operator speech into text. The real-time captured image from the camera and the text prompt are subsequently passed to the GPT-4o API to generate manipulation type recommendations, which are then processed to automatically switch or adjust the corresponding control mode or joint configuration.

\subsection{Type Adjustment Strategy}
Our system supports type adjustment to further enhance its versatility, while our type library can already cover most common tasks, and each type generalizes well across objects with similar geometric characteristics. To enable such adjustment, the system allows users to explicitly apply offsets to the position or orientation of specific fingertips.

Specifically, the system first obtains the initial fingertip position and orientation of the origin type through forward kinematics. And the desired adjustment can be specified either by capturing the 6-DOF motion of the user’s fingertip or by providing transformation values obtained through manual input. The system then applies the offset to the fingertip pose and uses the resulting position and orientation to compute the adjusted joint angles via inverse kinematics, formulated as:
\begin{equation}
q' = IK\left( FK(q) \cdot T_\Delta \right)
\end{equation}
where $q$ denotes the initial joint angles, $FK(\cdot)$ represents the forward kinematics function, $T_\Delta$ is the desired transformation applied to the end-effector pose, $IK(\cdot)$ denotes the inverse kinematics function; and $q'$ is the resulting new joint configuration. To ensure that the adjusted type remains as close as possible to the original pose, the system initializes the inverse kinematics solver with the joint angles of the origin type. This strategy helps avoid unintended deviations or discontinuities in the resulting posture.

\subsection{Interpolation Mapping Strategy}
We design an interpolation mapping strategy to intuitively control robotic dexterous hands using human hand motions. Specifically, we first associate the stretched and contracted postures of the human hand with the corresponding postures of the robotic hand. Given the current human hand posture, we compute a normalized projection ratio for each fingertip position along the 3D vector defined by the stretched and contracted positions:

\vspace{-1em}
\begin{equation}
p_{\text{ratio}} = \text{clip}\left(
\frac{
    (\mathbf{p}_{\text{current}} - \mathbf{p}_{\text{stretch}}) \boldsymbol{\cdot}
 (\mathbf{p}_{\text{contract}} - \mathbf{p}_{\text{stretch}})
}{
    \|\mathbf{p}_{\text{contract}} - \mathbf{p}_{\text{stretch}}\|^2},\, 0,\, 1\right),
\end{equation}

where \( \mathbf{p} \in \mathbb{R}^3 \) denotes the 3D fingertip position, \( \boldsymbol{\cdot} \) denotes the dot product, and \( \text{clip} \) constrains the output within the range \([0, 1]\). The resulting scalar \( p_{\text{ratio}} \) is then used to linearly interpolate between the stretched and contracted joint angles of the robotic hand:
% \vspace{-1em}
\begin{equation}
    \theta_{current} = p_{ratio}\boldsymbol{\cdot} (\theta_{contract}-\theta_{stretch})+\theta_{stretch},
\end{equation}

where \( \theta_{contract} \) and \( \theta_{stretch} \) represent the joint angles corresponding to the fully contracted and stretched states, respectively.

\subsection{Hardware and Robot Control}
\label{Hardware and Robot Control}
The hardware of our system involves the hand motion capture device, robot arm, dexterous hand and camera. For motion capture device in teleoperation, we employ Rokoko Gloves to capture 3 DOF position of each finger and employ the controller of Meta Quest 3 VR to capture wrist 6 DOF pose, following \cite{chen2024arcap}. For the robot, we employ two Kinova arms (6 DOF and 7 DOF, respectively) and two LEAP dexterous hands (16 DOF each) \cite{shaw2023leap}. For vision data collection, we employ a Realsense L515 LiDAR Camera to capture a single-view RGB-D observation of the scene. 

For dexterous hand control, we use joint position PD control, where the target position is obtained through interpolation in the mapping. For arm control, we utilize the high-frequency Cartesian velocity control \cite{ziegler1942pid} interface provided by Kinova. The arm's motion is predefined as a uniform acceleration and deceleration motion for smoothness. The maximum translational velocity is fixed at 20 cm/s. And the rotational velocity is dynamically adjusted: it increases when the orientation error is large, with an upper bound enforced to ensure safety. Additionally, Kalman filtering  \cite{kalman1960kalman} is applied to smooth the estimated velocity signals and further enhance the continuity of motion. More details of robot control can be found in supplementary materials. 

\begin{figure}[t]
    \centering
    \includegraphics[width=0.8\linewidth]{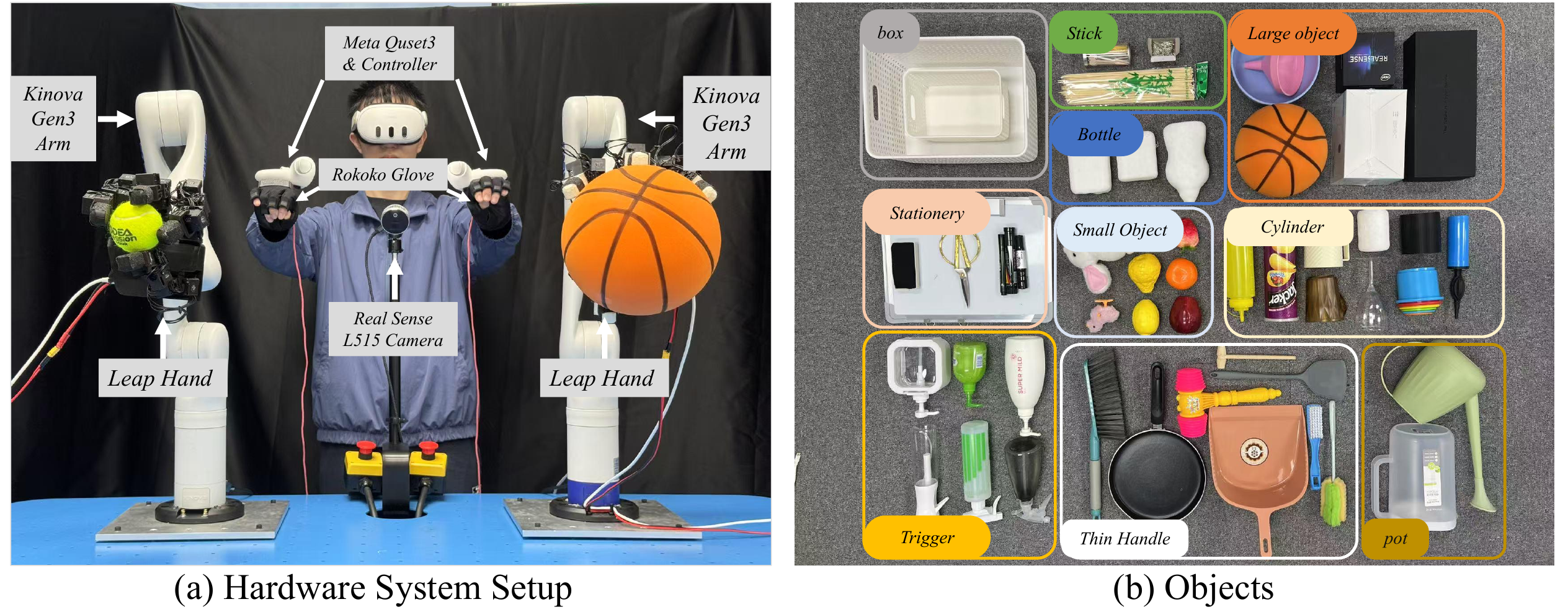}
    \caption{
    The illustration of hardware system setup and the objects used in experiments. 
    % \textcolor{red}{!!!!Warning: fig(b) seems to have not been finished yet}
    }
    \label{fig:realworldsetting}
\vspace{-1em}
\end{figure}

\section{Experiment}

\subsection{Experiment Setting and Evaluation Metrics}

\textbf{Tasks}. We design a diverse set of tasks to evaluate the effectiveness of both our teleoperation system and the imitation learning policy. The experiments cover both single-hand and bimanual manipulation. For better comparison, we referenced prior work in task design and adopted several existing tasks \cite{cheng2024Open-TeleVision, Chen2024BunnyVisionPro, chen2024arcap, yang2024ace}. Additionally, we introduce more challenging tasks that are difficult to complete using previous systems, in order to demonstrate the superior performance of our framework. Details of the tasks are provided in the supplementary materials. 

\textbf{Teleoperation Setting and Metrics}. We compare our type-guided teleoperation system with the retargeting-based baseline, where human hand postures are directly mapped to the robot \cite{qin2023anyteleop, chen2024arcap, Chen2024BunnyVisionPro}. Both systems share identical hardware and robot control algorithms to ensure a fair comparison. We conduct a user study involving 10 participants with varying levels of prior experience in robotics and teleoperation. Each participant is asked to perform two tasks, randomly selected from the task set. For each task, participants complete 20 successful demonstrations. We record the success rate $Suc$, the total time spent of completing the data collection of one task $T_{all}$ including the time for failure cases, and the average demonstration duration $T_{single}$ for each successful demonstration. Higher success rates and shorter durations indicate better performance \cite{Chen2024BunnyVisionPro}.

\textbf{Imitation Setting and Metrics}. We adopt the state-of-the-art diffusion-based policy, iDP3 \cite{ze2024idp3}, as our imitation learning algorithm. The policy takes a single-view 3D observation and current robot proprioception as conditional inputs, and outputs the desired Cartesian position of the robot arm's end-effector and the joint angles of the dexterous hand. Observation and action horizon vary from 3–8 and 13-8 respectively, depending on task length. To evaluate the impact of teleoperation quality, we train separate policies on datasets collected from the retargeting-based and our systems, using the same number of demonstrations and policy hyper-parameters. Higher task success rates achieved by a policy indicate higher-quality demonstrations and thus more effective data collection.

\begin{table}[]
\centering
\begin{tabular}{cclcccc}
\toprule
\multicolumn{1}{c}{Task} & \multicolumn{1}{c|}{Description} & System & $Suc$ & $T_{all}$ & $T_{single}$   \\ \midrule

\multirow{2}{*}{Pick and Place} & \multicolumn{1}{l|}{\multirow{2}{*}{\shortstack[l]{Pick up the tennis ball and place \\ it into the basket.}}}
   & Baseline & 95.2\%  & 579.6 & 8.28   \\
 & \multicolumn{1}{c|}{} & \textcolor{NavyBlue}{Ours} & \textcolor{NavyBlue}{100\%} & \textcolor{NavyBlue}{536.9} & \textcolor{NavyBlue}{7.67} \\

 \multirow{2}{*}{Collect and Store} & \multicolumn{1}{l|}{\multirow{2}{*}{\shortstack[l]{Collect objects on the table and \\ store them into a basket.}}} 
   & Baseline &60.6\%  & 1231.6 &  37.32   \\
 & \multicolumn{1}{c|}{} & \textcolor{NavyBlue}{Ours} & \textcolor{NavyBlue}{95.2\%}  & \textcolor{NavyBlue}{616.8} & \textcolor{NavyBlue}{29.37}   \\ 

 \multirow{2}{*}{Handover} & \multicolumn{1}{l|}{\multirow{2}{*}{\shortstack[l]{Transfer the object from the left \\ hand to the right hand.}}} 
   & Baseline &80.0\%  & 459.5 &  18.38   \\
 & \multicolumn{1}{c|}{} & \textcolor{NavyBlue}{Ours} & \textcolor{NavyBlue}{95.2\%} & \textcolor{NavyBlue}{244.4} & \textcolor{NavyBlue}{11.64}   \\ 
 
\multirow{2}{*}{Pouring from Pan} & \multicolumn{1}{l|}{\multirow{2}{*}{\shortstack[l]{Grasp the handle of pan and pour \\ its contents into the basket.}}} 
   & Baseline &14.2\%  & 1149.4 &  16.42   \\
 & \multicolumn{1}{c|}{} & \textcolor{NavyBlue}{Ours} & \textcolor{NavyBlue}{83.0\%}  & \textcolor{NavyBlue}{174.9} & \textcolor{NavyBlue}{14.57}   \\ 

   \multirow{2}{*}{Use Scissors} & \multicolumn{1}{l|}{\multirow{2}{*}{\shortstack[l]{Use scissors to cut the paper strip \\ into two pieces.}}} 
   & Baseline & 0  & - &  -  \\
 & \multicolumn{1}{c|}{} & \textcolor{NavyBlue}{Ours} &\textcolor{NavyBlue}{91.1\%}  & \textcolor{NavyBlue}{161.1} & \textcolor{NavyBlue}{5.37}   \\ 
 
 \multirow{2}{*}{Spray Water} & \multicolumn{1}{l|}{\multirow{2}{*}{\shortstack[l]{Use the spray bottle to spray \\ water toward the target direction.}}} 
   & Baseline & 0 & - &  -  \\
 & \multicolumn{1}{c|}{} & \textcolor{NavyBlue}{Ours} & \textcolor{NavyBlue}{86.9\%} & \textcolor{NavyBlue}{167.4} &  \textcolor{NavyBlue}{7.28}  \\

 \multirow{2}{*}{Use a \underline{\textbf{Heavy}} Kettle} & \multicolumn{1}{l|}{\multirow{2}{*}{\shortstack[l]{Grasp the kettle, lift it and \\ pour water into the container.}}} 
   & Baseline & 0  & - & -   \\
 & \multicolumn{1}{c|}{} & \textcolor{NavyBlue}{Ours} & \textcolor{NavyBlue}{85.0\%} & \textcolor{NavyBlue}{369.4} & \textcolor{NavyBlue}{18.47}   \\ 

  \multirow{2}{*}{Open a \underline{\textbf{Large}} Box} & \multicolumn{1}{l|}{\multirow{2}{*}{\shortstack[l]{Open the lid of the large box, \\ then pick up the items inside.}}} 
   & Baseline & 0  & - & -   \\
 & \multicolumn{1}{c|}{} & \textcolor{NavyBlue}{Ours} & \textcolor{NavyBlue}{95.2\%}  & \textcolor{NavyBlue}{398.6} &  \textcolor{NavyBlue}{18.98}  \\ 

  \multirow{2}{*}{Grasp Two Objects } & \multicolumn{1}{l|}{\multirow{2}{*}{\shortstack[l]{Grasp two medium-sized objects \\ by one hand simultaneously.}}} 
   & Baseline & 0 & - & -   \\
 & \multicolumn{1}{c|}{} & \textcolor{NavyBlue}{Ours} & \textcolor{NavyBlue}{69.6\%} & \textcolor{NavyBlue}{488.3} & \textcolor{NavyBlue}{21.23}   \\ 
 
\bottomrule
\end{tabular}
\vspace{1em}
\caption{The teleoperation results compared with retargeting-based teleoperation system (baseline).}
\vspace{-1em}
\label{table: teleoperation results}
\end{table}

\begin{figure}[t]
    \centering
    \includegraphics[width=\linewidth]{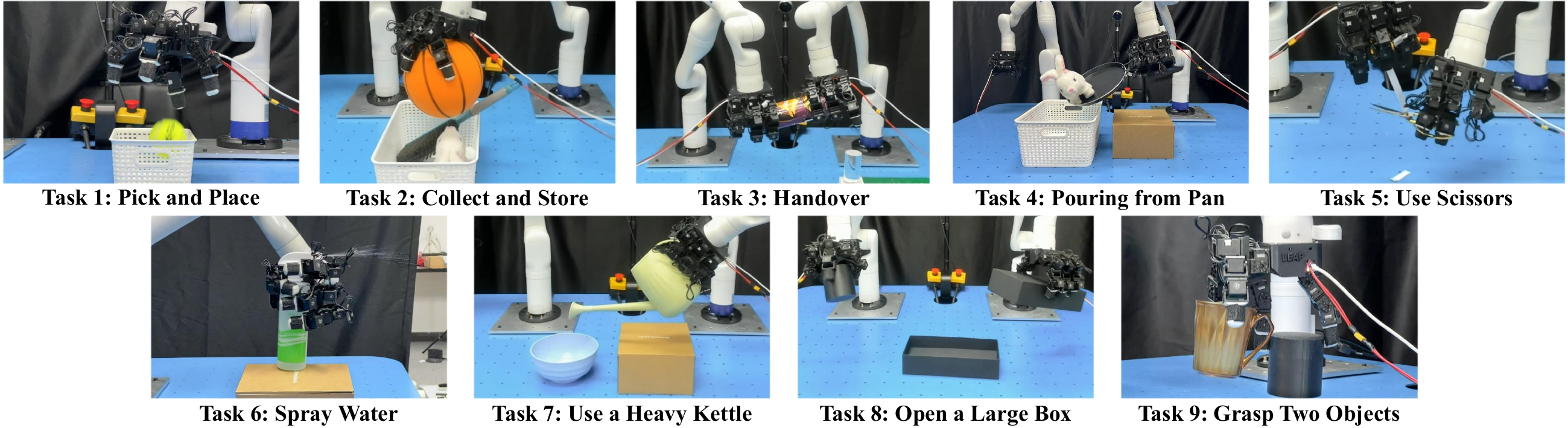}
    \caption{
    The visualization of autonomous policy execution process.
    }
    \label{fig:applicability}
\vspace{-1em}
\end{figure}

\subsection{Comparison Results}

\textbf{TypeTele significantly improves the efficiency of data collection.}
The teleoperation results in Table \ref{table: teleoperation results} show that our system achieves a shorter overall collection time, a higher task success rate, and reduced lengths of averaged demonstration trajectory lengths. This indicates that leveraging dexterous manipulation types enables more convenient object manipulation and more stable grasping, which is highly beneficial for improving teleoperation performance.

\textbf{TypeTele enables the successful execution of tasks that are unachievable using retargeting-based teleoperation system,} as shown in the results for challenging tasks (the final 5 tasks) presented in Table \ref{table: teleoperation results}. Specifically, when using scissors and spray bottles, the dexterous hand requires accurate manipulation while maintaining a stable grasp. For using a heavy kettle, a firm grip is necessary to counteract the object's weight. When opening a large box or grasping two objects, the hand must open widely to securely grasp and lift the lid or open to specific postures.
These tasks are particularly challenging for retargeting-based teleoperation systems, due to issues of unstable grasping and undesired contacting, as mentioned in Figure \ref{fig:problem}. These results highlight the enhanced capability of our system in handling complex and dexterous manipulation tasks.

\textbf{Imitation learning results demonstrate the higher quality of data collected by TypeTele.} As the quality of demonstrations affects the effectiveness of imitation learning, we compare the performance of an autonomous policy trained with an equal amount of successful demonstrations collected by different systems. The results shown in Table \ref{table: imitation} indicate that the policy trained with data collected by our system achieves a higher success rate, evaluated within 10 attempts. The tasks are ordered consistent with the tasks in Table \ref{table: teleoperation results}. The results with ``-'' are the tasks that can not be completed by baseline teleoperation. These results highlight the importance of high-quality demonstration data and the higher quality teleoperation of our system.

\begin{table}[t]
\centering
\begin{tabular}{>{\centering\arraybackslash}p{1.5cm}|cccccccccc|}
\toprule
\multicolumn{1}{c}{} & Task1 & Task2 & Task3 & Task4 & Task5 & Task6 & Task7 & Task8  & Task9 \\ \midrule
\multicolumn{1}{c|}{Baseline} &  10/10&  3/10&  1/10&  1/10&- & - &  -&   -& -  \\
\multicolumn{1}{c|}{Ours} &  10/10&  10/10&  6/10&  9/10& 9/10 & 9/10 &  9/10& 9/10 & 8/10\\  \bottomrule
\end{tabular}
\vspace{1em}
\caption{Comparison with the imitation policy trained using data collected by different teleoperation.}
\label{table: imitation}
\vspace{-1em}
\end{table}

\begin{figure}[t]
    \centering
    \includegraphics[width=\linewidth]{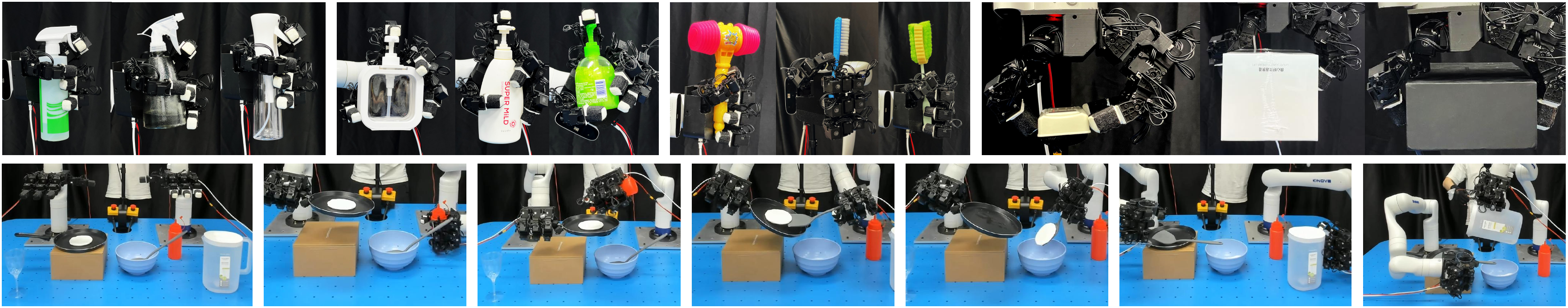}
    \caption{
    Top: Visualization that one type can apply to various objects with similar structures or functions. Bottom: Visualization of long horizon task involves mutlpile steps and objects.
    }
    \label{fig:applicability}
\vspace{-1em}
\end{figure}

\begin{figure}[t]
    \centering
    \includegraphics[width=\linewidth]{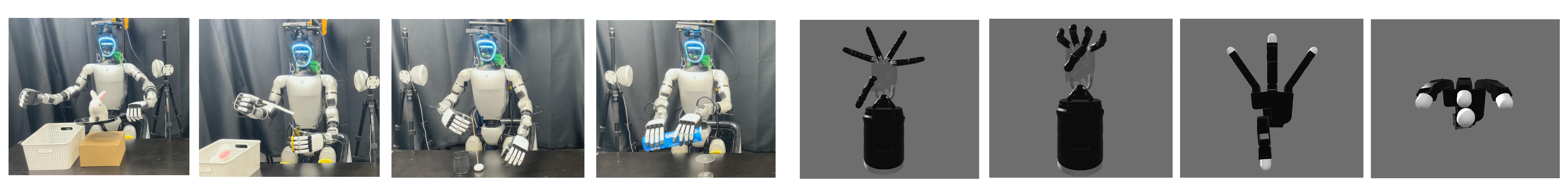}
    \caption{
    Left: Visualization of the experiments for Inspire Hand. Right: Visualization of types for Shadow Hand and Allergo Hand.
    }
    \label{fig:new hands}
\vspace{-1em}
\end{figure}

\subsection{Applicability of the TypeTele System}

\textbf{One dexterous manipulation type can be applied to various objects with similar geometric structures or functional properties.} As shown in the above of Figure \ref{fig:applicability}, the type designed for objects like trigger sprayers and lotion pumps can generalize across different instances. And the type designed for square-shaped and objects with thin handle can adapt to objects of varying sizes. These results demonstrate the broad applicability of our dexterous manipulation type design. 

\textbf{TypeTele can handle a wide range of complex manipulation tasks}, including long-horizon scenarios involving multiple objects and stages as shown in the bottom of Figure \ref{fig:applicability}. Our system is capable of selecting and executing the appropriate type at each stage, demonstrating strong adaptability and generalization. This enables TypeTele to successfully accomplish challenging tasks that involve diverse object interactions and multiple steps. 

\textbf{TypeTele is applicable to various dexterous robotic hands}, as illustrated in Figure~\ref{fig:new hands}. We conduct real-world manipulation experiments using the Inspire Hand, and evaluate type construction and grasp motion in simulation with both the Shadow Hand and the Allegro Hand.

\subsection{Efficiency of TypeTele System}
During teleoperation, the system records data at 15 FPS using a Windows 10 PC with an Intel Core i7-14700 CPU. Inference with the imitation policy runs at 11 FPS on an NVIDIA GeForce RTX 3090 GPU. An independent control thread for the robotic arm consistently maintains a frame rate of 25 FPS during both teleoperation and inference. For the MLLM-assisted retrieval module, the average query time over three trials is 4.8 seconds. While the retrieval step is relatively time-consuming, it occurs only once per task and thus has minimal impact on the system’s real-time performance. These results demonstrate that TypeTele maintains practical efficiency across all major system components.

\subsection{Effectiveness of MLLM-assisted Retrieval Module}
Experiments are conducted to evaluate the effectiveness of our MLLM-assisted retrieval module. We construct 50 test environments, including 40 single-object tasks and 10 multi-object long-horizon tasks. A retrieval is considered successful if the retrieved manipulation type is suitable for the current task. The success rates are 91.89\% for single-object environments and 92.00\% for multi-object tasks. This high accuracy confirms that our retrieval module can reliably identify appropriate manipulation types for diverse tasks.

\section{Conclusion}
\label{sec:conclusion}
We believe that achieving effective teleoperation for the data collection of delicate dexterous manipulation task is important in the robotic learning communities. In this paper, we propose Typetele, a novel dexterous teleoperation system with the insight that introducing types into teleoperation. To support this system, we build a dexterous manipulation library, comprising various types required for common dexterous tasks. During the teleoperation, a MLLM-assisted type retrieval module is proposed to select the suitable type for current task. And a interpolation mapping is used to control the dexterous hand by human hand motion. The extensive experiments show that our system not only enables tasks previously unachievable by teleoperation, but also greatly improves data collection efficiency and quality, thereby enhancing imitation learning and autonomous policy performance.

\bibliography{example}  % .bib

\newpage

\begin{center}
    \Large\textbf{Supplementary Materials}
\end{center}

\section{TypeTele System Details}

\subsection{Dexterous Manipulation Type Library}

\textbf{Visualization of Type Library}

\begin{figure}[h]
    % \vspace{-1em}
    \centering
    \includegraphics[width=\linewidth]{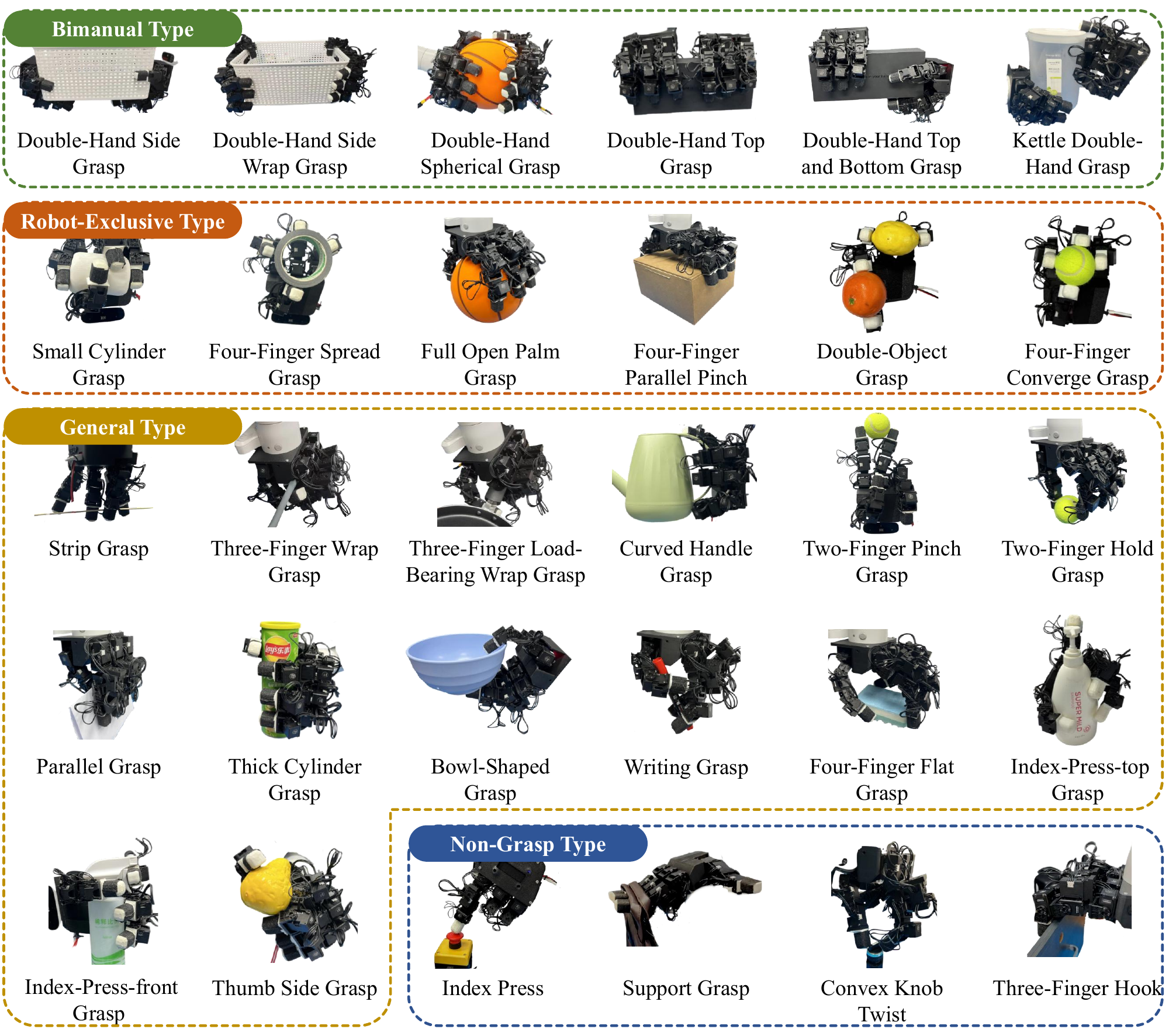}
    \caption{
    Visualization of Dexterous Manipulation Type Library.
    }
    \label{fig:typelib}
    % \vspace{-1em}
\end{figure}

We construct a dexterous manipulation type library using the taxonomy, based on prior grasp type work \cite{feix2015grasptaxonomy1, cutkosky1989grasptaxonomy2, krebs2022bimanualtaxonomy, fang2025anydexgrasp, chen2024non-grasp}, and extending it based on the structure of dexterous hands \cite{shaw2023leap} and manipulation tasks. The visualization of the library is illustrated in Figure \ref{fig:typelib}.

\textbf{Annotation Information of Dexterous Manipulation Types}

\begin{figure}[h]
    % \vspace{-1em}
    \centering
    \includegraphics[width=\linewidth]{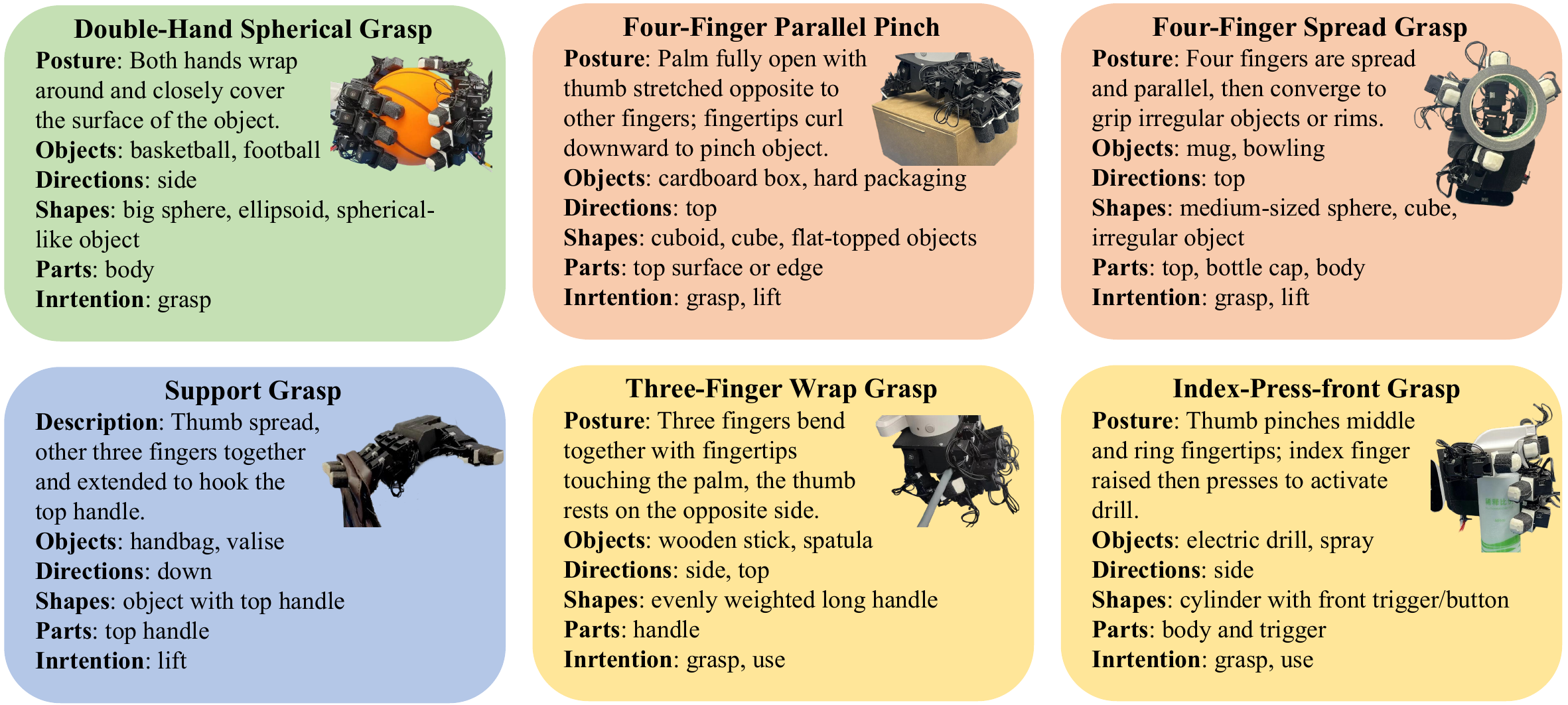}
    \caption{
    Visualization of the annotations examples of types in the library.
    }
    \label{fig:type_ano}
    % \vspace{-1em}
\end{figure}

Each manipulation type is annotated with descriptive information to characterize its posture and functionality, which facilitates retrieval. The annotated attributes include: \textit{hand posture}, \textit{manipulable object categories}, \textit{contact parts on the object}, \textit{the geometry of these parts}, \textit{grasp direction}, and \textit{intended manipulation purpose}. Examples of the annotated information are shown in Figure \ref{fig:type_ano}.

\subsection{MLLM-assisted Type Retrieval Module}
We employ GPT-4o \cite{hurst2024gpt} to retrieve the most suitable dexterous type for current task. The details of prompts are as following:

\begin{lstlisting}
System Prompt:
You are a robotic manipulation expert. When given a user goal and an image of tools or ingredients, your job is to: 
(1) Decompose the task into clear manipulation steps. 
(2) Assign a suitable grasping type for each hand (left/right) in every step based on the provided grasp type library. 
(3) Format your response in this structured way: 
The task is divided into N steps:  
Step 1: [describe the subtask]  
Step 2: [describe the subtask]  
... 
The types in each step are: 
Step 1:  Left type: [grasp type name]  Right type: [grasp type name]   
Step 2:  Left type: [grasp type name]  Right type: [grasp type name]  
... 
User Command: I want to [describe the whole task].
\end{lstlisting}

\textbf{Example of Output}

User Command
\begin{lstlisting}
I want to prepare a pancake with tomato sauce and a glass of water for breakfast.
\end{lstlisting}
GPT-4o
\begin{lstlisting}
The task is divided into 3 steps:
Step 1: Pick up the pan with the pancake using the right hand, and hold and squeeze the tomato sauce bottle using the left hand to apply sauce.
Step 2: Keep holding the pan with the right hand, and use a spatula with the left hand to transfer the pancake into the bowl.
Step 3: Hold the water pitcher with the left hand and the cup with the right hand, then pour water into the cup.
The types in each step are:
Step 1:
Left type: Thick Cylinder Grasp (for squeezing the sauce bottle)
Right type: Three-Finger Load-Bearing Wrap Grasp (for lifting the pan by the handle)
Step 2:
Left type: Three-Finger Wrap Grasp (for using the spatula)
Right type: Three-Finger Load-Bearing Wrap Grasp (continue holding the pan)
Step 3:
Left type: Curved Handle Grasp (for holding the pitcher)
Right type: Thick Cylinder Grasp (for holding the glass)

\end{lstlisting}

\subsection{Kinesthetic Teach Module}
Although the types provided in our type library can handle the most of everyday applications,  we offer a teaching mode for the creation of new type for special cases and unique user needs. This allows users without robotics expertise to intuitively and conveniently create dexterous types.  

We implemented the teaching mode using admittance control and motor backdrivability. The formula for admittance control is as follows:  
\begin{equation}
M\ddot{x}(t) + B\dot{x}(t) + Kx(t) = F_{\text{ext}}(t)
\end{equation}
In this equation, \(x(t)\) denotes the position of the control output, \(\dot{x}(t)\) and \(\ddot{x}(t)\) represent the velocity and acceleration respectively, and \(F_{\text{ext}}(t)\) is the external force applied to the robot's end-effector. The parameters \(M\), \(B\), and \(K\) correspond to the virtual mass, damping, and stiffness, respectively.

Here, we estimate the external force using the current magnitude and positional deviation of the dexterous hand's motors. We also incorporate the motor's velocity information to give the motion a certain degree of inertia, making the teaching process smoother.

\subsection{Robot Control}

To enhance the control fidelity and operational fluidity of the robotic arms during both teleoperation and imitation learning inference, we implemented several key methodological improvements: 

\textbf{(1) Multi-threaded Device Communication:} We have adopted a multi-threaded approach for device communication. Each distinct data stream – including RGB-D imagery, cartesian poses of the two end-effectors, and joint angles of the two robotic hands – is managed by an independent thread. This architecture ensures that when the main thread requires specific information, it can be provided instantaneously, thereby circumventing delays typically associated with data acquisition operations. 

\textbf{(2) Uniformly Accelerated Motion for Velocity Control:} For both translational and rotational velocity control, we have applied uniformly accelerated motion profiles. This strategy guarantees that velocity changes are smooth and devoid of abrupt transitions. Consequently, the robotic arm's movements are exceptionally fluid, and this approach also mitigates jitter stemming from natural human hand tremors or sensor inaccuracies. 

\textbf{(3) Dynamic Speed Control for Rotation:} A dynamic speed control scheme has been implemented for rotational movements. When the current orientation is significantly distant from the target orientation, the rotational speed will be increased, enabling the robotic arm to rapidly converge towards the desired direction. Conversely, as the current orientation approaches the target, the rotational speed will be reduced. This allows the operator to perform precise, fine-grained rotational adjustments. 

\textbf{(4) Dedicated Asynchronous Robot Control Thread:} The robotic arms are controlled by a dedicated, separate thread, ensuring asynchronous operation. The main thread focuses solely on transmitting the target pose to this robot control thread. Subsequently, the robot control thread governs the robotic arms at a consistent control frequency. This approach guarantees stable and smooth robotic arm control, even when the main thread's frame rate fluctuates or varies, such as during transitions between teleoperation and imitation learning inference with their differing frame rates.

\subsection{Details of Imitation Learning}
We adopt a diffusion-based imitation policy to learn from expert demonstration data, following \cite{ze2024idp3}. The observation input consists of single-view point clouds $x_i \in \mathbb{R}^{N \times 3}$ and robot proprioceptive inputs $x_p \in \mathbb{R}^{p}$. Specifically, we randomly downsample $N = \text{4096}$ points from the raw depth maps. The proprioceptive input ($p = \text{44}$) includes the Cartesian poses of both robot arms and the joint angles of the two dexterous hands.

The point clouds are encoded using a pyramid convolutional encoder \cite{ze2024idp3}, while the proprioceptive inputs are processed via a multilayer perceptron (MLP). We define the observation horizon as $t_o$ and the action horizon as $t_a$. In our setup, we adopt a fixed total horizon length of $t_o + t_a - 1 = 15$. For tasks requiring longer-term reasoning or delayed consequences, we use longer observation horizons (e.g., $t_o = 6$ or $8$) and shorter action horizons (e.g., $t_a = 10$ or $8$), while for reactive or short-horizon tasks, we opt for shorter observations (e.g., $t_o = 3$ or $4$) and correspondingly longer action horizons. This enables a flexible temporal encoding of task-relevant information, tailored to the nature of each behavior. All features are used as conditional inputs to predict the noise associated with the robot action $a \in \mathbb{R}^{k_a}$, where $k_a$ denotes the action dimension specific to the task. The training objective minimizes the denoising score matching loss, formulated as:
\begin{equation}
\mathcal{L} = \mathbb{E}_{\mathbf{a}_0, \boldsymbol{\epsilon} \sim \mathcal{N}(0, \mathbf{I})} \left[ \left\| \boldsymbol{\epsilon} - \epsilon_\theta(\mathbf{a}_t, t) \right\|^2 \right],
\end{equation}
where $\boldsymbol{\epsilon} \sim \mathcal{N}(0, \mathbf{I})$ is the Gaussian noise. The network $\epsilon_\theta$ is trained to predict the added noise given the noisy action $\mathbf{a}_t$ and the timestep $t$. We employ DDIM \cite{song2020ddim} for inference sampling.
\begin{equation}
\mathbf{a}_{t-1} = \sqrt{\bar\alpha_{t-1}} \left( \frac{\mathbf{a}_t - \sqrt{1 - \bar\alpha_t} \, \epsilon_\theta(\mathbf{a}_t, t)}{\sqrt{\bar\alpha_t}} \right) + \sqrt{1 - \bar\alpha_{t-1}} \cdot \epsilon_\theta(\mathbf{a}_t, t)
\end{equation}
where ${\bar\alpha}_{t-1}$ and $\bar{\alpha}_{t}$ are the cumulative noise schedule coefficients at time steps $t-1$ and $t$, respectively.

\section{Experiments Details}
\subsection{Details of Tasks}

\textbf{Task 1: Pick and Place.} Task 1 is a fundamental task that requires picking up a tennis ball on the table and placing it into a basket.

\textbf{Task 2: Collect and Store.} Task 2 focuses on the integrated capabilities of the system. Task 2 requires collecting three objects from the table and placing them into the basket in the following order: doll, broom, and basketball. For TypeTele, the operator can use voice commands to switch types for objects with different geometric shape during teleoperation.

\textbf{Task 3: Handover}. Task 3 evaluates the system’s bimanual coordination capabilities and grasp robustness. In this task, the left hand is required to pick up a can from a stand and then hand it over to the right hand.

\textbf{Task 4: Pouring from Pan.} Task 4 requires stably grasping a pan and then pouring the contents of the pan into the basket. The difficulty of this task lies in the need for the hand to firmly grip the pan's handle to prevent tilting or dropping during the pouring process.

\textbf{Task 5: Use Scissors.} In Task 5, the right hand is required to hold a strip of paper while the left hand uses a pair of scissors to cut through it. The task is considered successful if the lower part of the paper strip is completely severed in a single cut.

\textbf{Task 6: Spray Water.} Task 6 requires grasping a spray bottle and then pressing the trigger to spray water. The task is considered successful if a stream of water is sprayed out.

\textbf{Task 7: Use a Heavy Kettle.} Task 7 evaluates the ability to operate under extreme weight. Task 7 requires firmly gripping the handle of a watering kettle filled with water, lifting the watering can, and then pouring water into a bowl.

\textbf{Task 8: Opening a Large Box.} Task 8 evaluates the ability to manipulate objects of significant size. In this task, the left hand is used to open a large box, followed by the right hand retrieving the object contained within.

\textbf{Task 9: Grasp Two Objects.} Task 9 aims to fully leverage the dexterity of the dexterous hand. This task requires using one hand to grasp two objects, first a water cup and then a cylinder.

\subsection{Additional Experiments}

We conducted additional experiments to further evaluate our teleoperation system through a user study involving five participants with varying levels of prior teleoperation experience. Each participant was instructed to complete an identical task (grasping the handle of a frying pan) using both the TypeTele system and a retargeting-based baseline. In order to mitigate the influence of learning effects, three participants used the TypeTele first, while the remaining two began with the baseline, and none were informed which system was the TypeTele and which was the baseline. Each system was tested in five trials per participant, during which we recorded Success Rate and Average Time per Success. The Average Time per Success is calculated by dividing the total time spent across all trials by the number of successful trials. This metric reflects the average amount of time required to obtain a single successful execution, capturing both task efficiency and failure overhead.

After completing all the tests, each participant completed a questionnaire that evaluated both systems in four dimensions: accuracy, responsiveness, ease of use, and user confidence. Each dimension was rated on a scale of 0-10. We then computed the average score for each dimension across all participants for both systems. The results are as follows:

\begin{figure}[h]
    % \vspace{-1em}
    \centering
    \includegraphics[width=\linewidth]{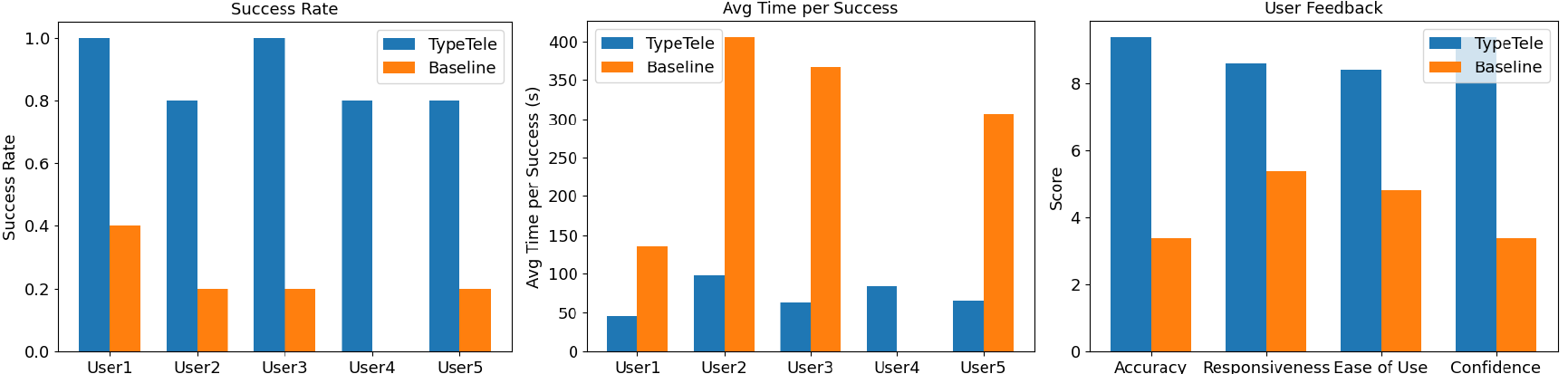}
    \caption{
    Results of User Study.
    }
    \label{fig:UserStudy}
    % \vspace{-1em}
\end{figure}

Experimental results demonstrate that the TypeTele system significantly outperforms the retargeting-based baseline across both objective and subjective measures. On average, TypeTele achieved a task success rate of 88\%, compared to only 20\% for the baseline. Participants also completed tasks faster using TypeTele. Subjective ratings further support these findings: TypeTele received higher scores across all four dimensions—accuracy (9.4 vs 3.4), responsiveness (8.6 vs 5.4), ease of use (8.4 vs 4.8), and user confidence (9.4 vs 3.4). These results indicate that TypeTele not only improves task performance but also delivers a more satisfying and trustworthy user experience.

\end{document}